\pdfoutput=1
\documentclass[a4paper]{article}

\usepackage{INTERSPEECH2022}
\usepackage{hyperref}

\title{Vakyansh: ASR Toolkit for Low Resource Indic languages}
\name{Harveen Singh Chadha$^1$, Anirudh Gupta$^1$, Priyanshi Shah$^1$, Neeraj Chhimwal$^1$, Ankur Dhuriya$^1$, Rishabh Gaur$^1$, Vivek Raghavan$^2$}
\address{
  $^1$Thoughtworks \\ $^2$Ekstep Foundation}
\email{\ harveen.chadha@thoughtworks.com,  vivek@ekstep.org}

\begin{document}

\maketitle
\begin{abstract}
  We present Vakyansh, an end to end toolkit for Speech Recognition in Indic languages. India is home to almost 121 languages and around 125 crore speakers. Yet most of the languages are low resource in terms of data and pretrained models. Through Vakyansh, we introduce automatic data pipelines for data creation, model training, model evaluation and deployment. We create 14,000 hours of speech data in 23 Indic languages and train wav2vec 2.0 based pretrained models. These pretrained models are then finetuned to create state of the art speech recognition models for 18 Indic languages which are followed by language models and punctuation restoration models. We open source all these resources with a mission that this will inspire the speech community to develop speech first applications using our ASR models in Indic languages.

\end{abstract}
\noindent\textbf{Index Terms}: Automatic Speech Recognition, Low Resource Languages, Indic Languages, Open Source

\section{Introduction}
End to End (E2E) systems for Automatic Speech Recognition (ASR) have gained popularity over the recent years. A great challenge lies for low resource languages where getting aligned acoustic and text data is difficult. Thus, getting good quality data becomes as important as using a proper architecture for model training. Furthermore, a language model (LM) can have a significant impact in improving the ASR results, provided the text corpus for LM creation was diverse enough.

For Indic languages, there is still a lot of gap in terms of open source speech toolkits containing end to end tools for training and deployment of ASR models. We present Vakyansh, developed specifically for Indic languages. Our contributions can be summarized as:

\begin{itemize}
    \item ASR models in 18 Indic Languages\footnote{\url{https://github.com/Open-Speech-EkStep/vakyansh-models}}.
    \item 14,000 hours of unlabelled data in 23 Indic languages.
    \item 10,000 hours of labelled data in 18 Indic languages.
    \item Data pipeline with all the components like: data discovery, data cleaning, transcript generation, forced alignment, chunking. 
    \item Post processing utilities for ASR: Punctuation restoration and Inverse Text Normalization
\end{itemize}

\section{Dataset}
The biggest challenge for creating ASR models for low resource Indic languages is the absence of data in open domain. We solve this problem by creating our own datasets in automated and semi automated manner for different languages. We started with Hindi which is spoken by 43 percent of the population in India and set a target of collecting 10,000 hours of labelled dataset which is required for any production grade ASR system.

We collect data through 3 different ways. First, transcript generation pipeline for languages where an ASR model is already present to collect dataset in automated manner. Second, forced alignment of text and audio where we have both the text and audio present. Third, where we just have the audio and we get that audio annotated through language experts. 

\subsection{What makes good data for ASR?}
The data collection process can be overwhelming if the definition of a good dataset is not defined clearly. We lay down a set of following rules for the properties of the desired dataset. 
\begin{enumerate}
    \item Audio data should come from different sources/domains.
    \item Break down longer audios into smaller chunks with duration ranging between 1 to 15 seconds.
    \item Contribution of one speaker in the dataset should not exceed a threshold of 90 minutes. This is to avoid model overfitting to a single speaker.
    \item Audios should have less background noise and music.
    \item Dataset should be balanced in terms of gender so as to not include any gender bias.
    \item Audio Data should be in the same language in which the ASR is desired. 
    \item The labelling process is such that it induces less noise to the transcripts.
    
\end{enumerate}

\subsection{Audio Processing Pipeline}
We create an Audio Processing pipeline which accomplishes all of the rules defined by us. We gave preference to audios present in the open domain.  This pipeline is used in the scenario where we have the audios of large duration and we want to convert them to smaller chunks.

\subsubsection{Data Discovery}
We create a scraper that scrapes the internet for open audios present. We start by browsing youtube for videos present in the open domain. We make sure that we collect data from different sources so that our training data comes from different domains.

\subsubsection{Voice Activity Detection (VAD)}
After the collection of Audios the first step is to break longer audios into smaller chunks. This is done to ensure manageable sized chunks fit easily to the GPU memory during training process. One technique is to break the audio every 10 seconds but it is really inefficient to do so because some words may break at the boundaries of the clipped audio and hence induce noise during the training or transcription process.

We ensure that words are not clipped by breaking the audios on silences by using WebRTC-VAD\cite{webrtc-vad}. We use an Aggressiveness level of 2 for our purpose. The WebRTC-VAD uses a padded, sliding window algorithm over the audio frames where if more than 90 percent of the frames in the window are voiced then the algorithm starts collecting audio frames. The collector waits until 90 percent of the frames in the window are unvoiced to detrigger and start collecting silence frames. We use a frame duration of 30ms and padding duration of 300ms.

\subsubsection{Signal to Noise Ratio (SNR)}
After the longer audio is broken into chunks, the first step is to remove individual chunks that contain a lot of background noise or music. We used WADA SNR \cite{WADA-SNR} to calculate the Signal to Noise Ratio. This approach assumes that the clean speech can be approximated by a Gamma distribution and additive noise signal is also Gaussian.

The chunks run through WADA SNR and we use this number as quality metric of the chunk. In our experimentation we found out that chunks with value less than 20 contain a lot of music and background noise. We set a minimum threshold of 20 SNR and maximum threshold of 60 SNR. We reject chunks having values outside the threshold range. We observe that 35 percent of our generated data gets rejected at this stage as it is classified as noise.

\subsection{Audio Metadata Processing}
To get a better sense of data distribution, we try to estimate the number of speakers and gender distributions in a given source.

\subsubsection{Speaker Clustering}
We assume that different sources do not have overlap of speakers, so we run a clustering algorithm source by source to filter out the 90 minutes of data for a particular speaker from a cluster. We make sure we select the best data for a particular speaker by sorting the chunks on SNR values and picking the top chunks which sum up to 90 minutes.

We use Resemblyzer\cite{Resemblyzer} to derive a 256 length embedding of the voice present in a chunk. These embeddings are then passed on to the HDBSCAN clustering algorithm where the clusters of different utterances are predicted. We then set a minimum utterance count threshold (required to create a cluster) to get a better estimate of the speakers present. We try this algorithm on a balanced dataset of 20 hours with 80 speakers in Hindi and found out that HDBSCAN predicts 79 speakers with almost $96\%$ accuracy. The accuracy here is defined by Cluster Purity which is the number of utterances belonging to the dominant speaker divided by total number of utterances in that cluster.  This approach gives us a rough sense of the speakers in the training data and also helps in filtering maximum data from a particular speaker.

\subsubsection{Gender Identification}
We make sure that our generated data contains sufficient chunks from both the genders. We run a classification algorithm which can give us an estimate of the balance of gender ratio in a dataset. To create data for modelling this problem, we use the same 256 embedding vectors as created in 2.3.1 as features for the classification problem. During the data preparation phase, we make sure that the dataset is balanced and comes from different languages like Hindi, Tamil, Telugu and Kannada.

A support vector machine with radial basis function  with $\gamma=0.01$ and $C=100$ worked best for this problem. The training data was of 34 hours with test set containing data of 8 hours from different languages and the accuracy on this test dataset was around 97 percent. When running on a cluster level, we made sure we picked the dominant gender as reported by the different utterances.

\begin{table}[th]
  \caption{Sources list}
  \label{tab:example2}
  \centering
  \scriptsize
  \begin{tabular}{ r@{}l  r }
    \toprule
    \multicolumn{2}{c}{\textbf{Source}} & 
                                         \multicolumn{1}{c}{\textbf{Code}} \\
    \midrule
    &Pipeline &PP \\
    &Forced Alignment &FA \\
    &Hand Labelled as &HL \\
    &Open SLR &OS \\
    &IITM English ASR Challenge &EN \\
    &IITM Indian ASR Challenge &IN \\
    &IIT Bombay Vaksancayah\cite{adiga2021automatic} &VA \\
    &MUCS 2021\cite{diwan2021multilingual} &MU \\
    &Common Voice\cite{CommonVoice} &CV \\
    \bottomrule
  \end{tabular}
  
\end{table}

\subsection{Transcription}
After processing the raw audio through 2.2 and 2.3 we now have the filtered audio chunks with their metadata. We pass these filtered audio chunks to a commercial Speech to Text engines to get them labelled. We make sure we use different variety of STTs, so that even if the noise is induced by a single STT, we make sure there are different types of noise. Also we set a hypothesis that if we use data from a single source then we can't create a better model than them, because our model will perform the same mistakes as theirs.

For very low resource languages, even a commercial Speech to Text Engine is not present. In such cases, we pass the same audio chunks to language experts for hand labelling the data. Table 2 contains all the information on how data for a particular language was created. Code for all the modules in the pipeline is present here.\footnote{\url{https://github.com/Open-Speech-EkStep/audio-to-speech-pipeline}}

\subsection{Forced Alignment Pipeline}
Apart from creating data from pipeline we also had some data where we had a large audio files and corresponding unaligned transcript files. This data came for News on Air bulletins. Audio recordings of the news were extracted and the corresponding transcript was extracted from PDF bulletins which could then be further used for forced alignment. 

The text was converted from \textit{KrutiDev} encoding to UTF-8 format. Audio segments from text were created using \texttt{espeak} TTS engine and Dynamic Time Warping (DTW) algorithm is used to align the Mel-frequency cepstral coefficients (MFCCs) representation of the given audio wave and the audio wave obtained by synthesizing the text fragments. Data generated through this procedure had some noise because the transcripts usually had words which were never spoken by the speaker so there were some alignment issues at the starting or ending of the sentences.

\subsection{Text Postprocessing Pipeline}
\subsubsection{Text cleaning}
We make sure that we remove all the punctuations and special characters from the transcripts. Only special characters which do not have a pronunciation in spoken speech are removed.

\subsubsection{Vocabulary Definition}
With the help of language experts we define the vocabulary of characters present in a particular language. Once this dictionary is defined, any other character appearing which is not present in the dictionary is considered as a foreign character. We run through all the transcripts generated by pipeline or language experts and remove utterances having foreign characters. We further loose about 5-10 percent data at this stage.

\subsubsection{Numbers deletion}
We decide to remove all the numbers as well from the vocabulary and any utterances containing numbers is also deleted. This was done to make sure that the trained model outputs the numbers in words rather than numbers because it is very difficult for model to learn different combinations of the digits if enough data for that is not present in the training.

Let's say, if there is data about currency and the transcripts contains only words like million then the model would never be able to understand hundred-thousands, billions because that text is never seen during the training. One solution to this problem is to convert all numbers to words using  num2words\cite{num2words} package but the problem here is that all these packages are not context dependent. For 4 digit numbers, when you say "I have 5000 dollars" then 5,000 is converted to five thousand. The similar conversion might not be good for other sentence like "The war started in 2022", here 2022 should be twenty twenty two and not two thousand twenty two. So its very hard to make a judgement of what was exactly said in the audio by the speaker.

\subsubsection{Text Normalization}
We also normalize the training text from \textit{Normalization Format C(NFC)} to \textit{Normalization Format D(NFD)}. The difference between both the normalization is shown in Figure 1. In NFD, a character is in decomposed form. We used Indic-NLP-library\cite{kunchukuttan2020indicnlp} to perform this NFC to NFD normalization for Indic languages. This helped us to reduce the vocabulary size of ASR and also make ASR work better for rare symbols that are combination of two or more characters.

\begin{figure}[t]
	\centerline{\includegraphics[scale=0.25]{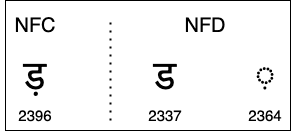}}
	\caption{NFC and NFD Format}
	\label{fig:Sample}
\end{figure}


\begin{table}[th]\centering
	\caption{Dataset Description}
	\label{tab1:hourly_desc}
	\centering
	\scriptsize
	\begin{tabular}{lrrrrr}
	    \toprule
		Language &Source &\multicolumn{2}{c}{Duration (hrs)} &Vocab Size\\
		\cmidrule{3-4}  
		& &Train &Dev & & \\
		\midrule
		Hindi &PP &4280 &17 &62  \\
		English &PP,IN &700 &10 &27 \\
		Kannada &PP &564 &10 &61\\
		Tamil &PP,IN,CV &255 &5 &48\\
		Bengali &OS &200 &10 &60 \\
		Nepali &OS &130 &7 &61\\
		Telugu &PP,MU &96 &5 &63\\
		Gujarati &PP,MU &96 &5 &64\\
		Odia &MU &95 &5 &60\\
		Marathi &MU &94 &5 &61\\
		Urdu &PP &60 &4 &43\\
		Bhojpuri &FA &56 &5 &60\\
		Sanskrit &VA &56 &7 &62\\
		Dogri &FA &51 &4 &64\\
		Maithili &FA &56 &3 &62\\
		Punjabi &PP,HL &8 &1 &60\\
		Malayalam &HL &7 &1 &67\\
		Assamese &HL &7 &1 &60 \\
		\bottomrule
	\end{tabular}
	
\end{table}

\subsection{Split}
While making a train/valid/test split, we make sure that the same speaker is not present across different sets. This was to avoid any leakage between different sets and get a fair estimate of the model quality on speakers not seen during the training time. The valid set is created carefully, mostly hand labelled in most of the cases, since we evaluate the performance of our model on valid set during training. 

\section{Methodology}
We create our own experimentation platform\footnote{\url{https://github.com/Open-Speech-EkStep/vakyansh-wav2vec2-experimentation}} on top of fairseq\cite{ott2019fairseq} as to experiment faster. We use wandb for experiment tracking. All our models are based on wav2vec 2.0\cite{baevski2020wav2vec}. We perform two types of training: pretraining and finetuning for ASR models. The ASR output is then combined with a KenLM\cite{heafield2011kenlm} based statistical language model to correct word spellings and improve performance of ASR. Metric for evaluation of ASR is WER (Word Error Rate) and CER (Character Error Rate).

\subsection{Pretraining}
The pretraining method we use masks the speech input in the latent space and solves a contrastive task defined over a quantization of the latent representations which are jointly learned and shows that powerful representations can be learnt from raw speech audio alone. \cite{baevski2020wav2vec} 
 
We use the base architecture of the wav2vec 2.0. It has \textit{12} blocks with a model dimension of \textit{768} and \textit{8} attention blocks. The pretraining is restored from a checkpoint which is trained on 960 hours of librispeech data in english. We choose the base architecture over large since it is faster to train and also has a much lesser inference time giving an edge in production systems.

For pretraining, only the audio files are needed and several chunks are concatenated together to create audios in the range of 25s-30s. Then we randomly crop audio samples at \textit{250,000} audio frames or \textit{15.6} seconds and use a dropout of \textit{0.1}. The model is trained for almost \textit{300,000} steps and start with a learning rate of \textit{0.0005}. We optimize using Adam where the first \textit{32,000} steps are used as warmup updates for the learning rate after which it is linearly decayed. A weight of $\alpha=0.1$ is used for diversity loss. We use $G = 2$ codebooks with $V = 320$ entries each for the quantization module. 

The experiments for pretraining are listed in Table 3. First, We train a hindi only monolingual model on 4200 hours, then we train a Kannada monolingual model on 1500 hours. Finally, we train a multilingual CLSRIL-23\cite{gupta2021clsril} model on 10,000 hours of data from 23 Indic languages. It took around 150 hours to reach a stage where we did not see any gain in code perplexity. 

\begin{table}[th]
  \caption{Pretraining Experiments and setup}
  \label{tab:example}
  \centering
  \scriptsize
  \begin{tabular}{lrrr}
    \toprule
    Language  &Model Type &Hours &Hardware \\
    \midrule
    Hindi &Monolingual &4200 &8 x V-100\\
    Kannada &Monolingual &1500 &8 x V-100\\
    CLSRIL-23 &Multilingual &10,000 &16 x A-100\\
    \bottomrule
  \end{tabular}
  
\end{table}

\subsection{Finetuning}
Finetuning is done on the labelled data. Pretrained models are fine-tuned by adding a fully connected layer on top of the context network with the size of output layer equal to the vocabulary of the language. Models are optimized using a CTC loss \cite{Graves06connectionisttemporal}. We finetune until we get the lowest WER on the valid set. 

We start with finetuning for Hindi on hindi pretrained model. We then create several models for other languages based on hindi pretrained model. We found that for these models to work it doesn't matters if the finetuned language is present in pretraining or not which meant that you can now create speech recognition models for low resource languages based on learnt representation for a high resource language. We then finetune on languages based on the multilingual CLSRIL-23 pretrained model. We present our results on 
CLSRIL-23 pretrained model in Table 4. Finetuning for all languages is performed on 8 Tesla V-100 GPUs. Hyperparameters for training are logged in wandb.

\subsection{Language Model}
The output from speech recognition model is fed to a statistical KenLM based language model where a beam search is performed to find the appropriate words and correct spellings of some common words. This further helps to improve the score by $20\%$ on average given the language model is general enough and is trained on large corpus of data.

We use IndicCorp\cite{indiccorp} to train most of our 5-gram KenLM based language models. The corpus for LM training is cleaned of all the characters/words that do not appear in the vocabulary of ASR training. The words are sorted based on their frequency and the top 5,00,000 words are picked. The probabilities are then calculated using 1-gram, 2-gram upto 5-gram. We use a beam search of \textit{128} for all the statistics reported in Table 4. We keep a word insertion penalty of -1 and LM weight of 2 which is basically a ratio of how much weightage is to be given to a language model as compared to original ASR output.

The application of LM can also increase the WER in some cases where the test set text is completely different from the LM training text. We also found out from our experiments that in order to create a Domain Specific ASR, it is very important for the LM to be trained on Domain specific text.

\section{Results}
We prefer to test on the datasets present in the open domain. We also make sure we do perform a in-domain and out-of-domain testing both in streaming and batch mode. Results of testing models in batch mode is available in Table 4.

\begin{table}[!htp]\centering
	
	\scriptsize
	\caption{Finetuning Results on 18 Indic languages}\label{tab1: exp2}
	\begin{tabular}{l|r|r|r|r|r}\toprule
		Language  &\multicolumn{2}{c}{Without LM}  &\multicolumn{2}{c}{With LM} &Test Set \\
		\cmidrule(lr){2-3} \cmidrule(lr){4-5}  
		&WER &CER &WER &CER  \\
		\midrule
		Hindi  &17.81 &5.74 &11.53 &4.22 &MU \\
		English &18.25 &6.73 &8.82 &4.4 &EN(5h) \\
		Kannada &29.06 &4.05 &26.94 &5.55 &OS\\
		Tamil &33.05 &6.76 &14.7 &3.79 &IN(5h)  \\
		Bengali &40.63 &13.14 &29.91 &12.47 &HL(5h) \\
		Nepali &15.71 &3.0 &9.4 &2.16 &OS \\
		Telugu &36.76 &8.96 &20.6 &5.11 &MU \\
		Gujarati &32.25 &11.17 &18.64 &7.89 &MU \\
		Odia &29.16 &5.24 &24.33 &4.07 &MU \\
		Marathi &24.6 &6.1 &14.0 &5.23 &MU \\
		Urdu &30.55 &17.01 &23.19 &18.86 &HL(10h) \\
		Bhojpuri &34.89 &25.11 &33.27 &27.61 &HL(5h)\\
		Sanskrit &19.15 &2.25 &45.1 &9.5 &VA\\
		Dogri &40.55 &23.7 &36.56 &24.8 &HL(3.5h)\\
		Maithili &12.8 &4.69 &12.24 &5 &HL(3h)\\
		Punjabi &28.32 &10.8 &23.93 &10.86 &HL\\
		Malayalam &59.77 &19.47 &47.48 &17.49 &HL(30m)\\
		Assamese &28.48 &7.3 &23.6 &6.6 &HL(1.2h) \\
		\bottomrule
	\end{tabular}
	
\end{table}

\subsection{What after ASR?}
The output of ASR is raw as this point as it doesn't include punctuations or numbers.

\subsubsection{Punctuation Restoration Models}
We use IndicCorp for training punctuation restoration models since before training the ASR we remove the punctuations. We pose punctuation restoration as a token classification task and use IndicBERT\cite{kakwani2020indicnlpsuite} for that downstream task. IndicBERT is a multilingual ALBERT \cite{lan2019albert} model trained on $12$ major Indian languages. It also has less parameters than BERT or XLM-R making it more suitable for the use of real time usage. 

\subsubsection{Inverse Text Normalization (ITN)}
ITN is the task of converting the raw output of spoken numbers by ASR model into its written form to improve text readability. We develop rule based WFST (Weighted finite state transducer) to achieve the same.

\section{Conclusions}

Vakyansh, through Open source contributions is an initiative to give a head start to researchers in Indic community to develop speech apps in local languages. This is our effort to push state of the art research in Indic Speech and create resources for low resource Indic languages so that other researchers can build upon it.

\section{Acknowledgements}

All of the authors are grateful to the Ekstep Foundation for financially supporting and providing the infrastructure for this research. Dr. Vivek Raghavan deserves special recognition for his constant support, mentoring, and constructive discussions. We also thank Nikita Tiwari, Anshul Gautam, Ankit Katiyar, Heera Ballabh, Niresh Kumar R, Sreejith V, Soujyo Sen and Amulya Ahuja for helping out when needed and infrastructure support for data processing, model training and model testing. \\

\bibliographystyle{IEEEtran}

\bibliography{vakyansh}


\end{document}